\documentclass[letterpaper, 10 pt, journal, twoside]{IEEEtran}
\usepackage{amsmath,amsfonts}
\usepackage{algorithmic}
\usepackage{algorithm}
\usepackage{array}
\usepackage{textcomp}
\usepackage{stfloats}
\usepackage{url}
\usepackage{verbatim}
\usepackage{graphicx}
\usepackage{cite}
\usepackage{color,soul}
\usepackage{caption}
\usepackage{subcaption}
\usepackage{pifont}
\usepackage{multirow}
\usepackage{siunitx}

\usepackage{booktabs}
\usepackage{tabularx}
\usepackage{colortbl}
\usepackage{diagbox}
\usepackage{makecell}
\usepackage{xcolor}
\usepackage{hyperref}
\definecolor{myred}{rgb}{0.796, 0.255, 0.329}

\hypersetup{
    colorlinks=true,
    linkcolor=blue,
    citecolor=blue
    }

\newcommand{\revtext}[1]{\textcolor{black}{#1}}

\begin{document}

\title{
Long-Term Upper-Limb Prosthesis Myocontrol \\ via
High-Density sEMG and Incremental Learning}

\author{Dario Di Domenico$^{1,2,\ddag}$, Nicolò Boccardo$^{1,3,\ddag}$, Andrea Marinelli$^{1,\ddag}$, Michele Canepa$^{1,3,\ddag}$,\\Emanuele Gruppioni$^{1,4,\ddag}$, Matteo Laffranchi$^{1,\ddag}$, Raffaello Camoriano$^{5,1}$

\thanks{This work was partially supported by INAIL (Istituto Nazionale per l'Assicurazione contro gli Infortuni sul Lavoro) under Grant Agreement: PR19-PAS-P1 - iHannes and PR23-PAS-P1 - Dexter Hand.
The Open University Affiliated Research Centre at Istituto Italiano di Tecnologia (ARC@IIT) is part of the Open University, Milton Keynes MK7 6AA, United Kingdom.
R. C. acknowledges the following: This study was carried out within the FAIR - Future Artificial Intelligence Research and received funding from the European Union Next-GenerationEU (PIANO NAZIONALE DI RIPRESA E RESILIENZA (PNRR) – MISSIONE 4 COMPONENTE 2, INVESTIMENTO 1.3 – D.D. 1555 11/10/2022, PE00000013). This manuscript reflects only the authors’ views and opinions, neither the European Union nor the European Commission can be considered responsible for them. }
        
\thanks{$^{\ddag}$Member IEEE}

\thanks{$^{1}$Dario Di Domenico, Nicolò Boccardo, Andrea Marinelli, Michele Canepa, Matteo Laffranchi and Raffaello Camoriano are with Rehab Technologies Lab, Istituto Italiano di Tecnologia (IIT), Genoa, Italy {(Corresponding author: \tt\footnotesize dario.didomenico@iit.it}, e-mail: {\tt\footnotesize name.surname@iit.it})}%
\thanks{$^{2} $Dario Di Domenico is with DET, Politecnico di Torino, Turin, Italy}%

\thanks{$^{3} $Nicolò Boccardo and Michele Canepa are with Open University Affiliated Research Centre (ARC@IIT), Genova, Italy}%

\thanks{$^{4} $Emanuele Gruppioni is with INAIL Prosthetic Center, Vigorso di Budrio, Italy
        {\tt\footnotesize e.gruppioni@inail.it}}%
        
\thanks{$^{5} $Raffaello Camoriano is with DAUIN, Politecnico di Torino, Turin, Italy}%

\thanks{Digital Object Identifier (DOI): see top of this page.}

}



\maketitle

\begin{abstract}

Noninvasive human-machine interfaces such as surface electromyography (sEMG) have long been employed for controlling robotic prostheses. However, classical controllers are limited to few degrees of freedom (DoF). More recently, machine learning methods have been proposed to learn personalized controllers from user data. While promising, they often suffer from distribution shift during long-term usage, requiring costly model re-training. Moreover, most prosthetic sEMG sensors have low spatial density, which limits accuracy and the number of controllable motions. In this work, we address both challenges by introducing a novel myoelectric prosthetic system integrating a  high density-sEMG (HD-sEMG) setup and incremental learning methods to accurately  control 7 motions of the Hannes 
prosthesis. First, we present a newly designed, compact HD-sEMG interface equipped with 64 dry electrodes positioned over the forearm. Then, we introduce an efficient incremental learning system enabling model adaptation on a stream of data. We thoroughly analyze multiple learning algorithms across 7 subjects, including \revtext{one with limb absence}, and 6 sessions held in different days covering an extended period of several months. The size and time span of the collected data represent a relevant contribution for studying long-term myocontrol performance. Therefore, we release the DELTA dataset together with our experimental code.
\end{abstract}


\vspace{0.7cm}
\section{Introduction}
\IEEEPARstart{H}{uman} interaction with the surrounding world is largely achieved through the execution of dexterous hand motions requiring synergical coordination between
forearm muscles to govern fingers and wrist movements. 
The loss of an upper limb severely affects an individual's life,
resulting in both social and psychological challenges.
Advancements in prosthetics and robotics can represent a source of empowerment for subjects facing such challenges, as they enable the recovery
of crucial functions, thus fostering greater independence.
Over the past decade, upper limb prostheses
have undergone significant enhancements, 
developing into
actual
robotic devices with multiple Degrees of Freedom (DoFs)~\cite{trent_narrative_2019}.
Concurrently, advancements in myoelectric control (myocontrol) have given rise to innovative approaches for the actuation of such complex devices, relying on electromyography (EMG) of residual muscle activity to encode the desired motions \cite{oskoei_myoelectric_2007}.

\begin{figure}[t]
    \centering
    \includegraphics[width=0.9\linewidth]{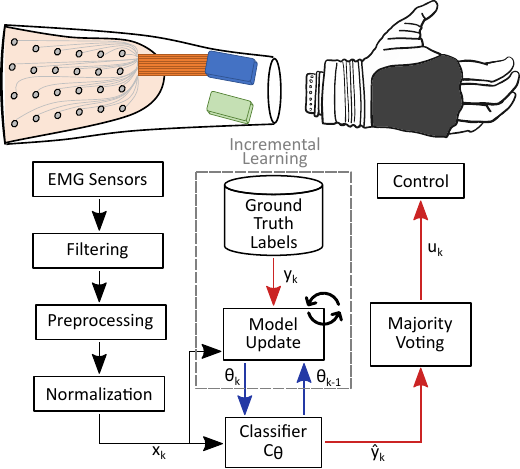} 
    \caption{Proposed system scheme.
    Pre-processed HD-sEMG signals $x_k$ at step $k$ are fed to the classifier $C_{\theta}$,
    which predicts the prosthesis gesture $\hat{y}_k$.
    $\hat{y}_k$ is filtered via time-window majority voting to improve control robustness.
    The dashed box shows the incremental model update process.}
    \label{fig:control_scheme}
\end{figure}
Nowadays, most available prosthetic solutions 
still rely on dual-site muscle contractions to trigger device movements, which in principle allow direct control of two actions only.
If more actions are available, the user can manually select the desired grasp or gesture configuration among several pre-recorded ones through remote control, apps, or physical buttons.
This results in considerable delays and unnaturalness in daily tasks, leading to dissatisfaction and diminished trust in the prosthesis~\cite{biddiss_upper-limb_2007}.
Prosthetic control strategies
also include threshold-based on/off or proportional control and pattern recognition methods~\cite{marinelli_active_2023}.
With the purpose of alleviating cognitive burden and 
enabling intuitive and precise 
myocontrol, researchers are also investigating the application of machine learning (ML) to predict 
desired hand motions~\cite{oskoei_myoelectric_2007, scheme_electromyogram_2011}.
Despite these advancements, learning-based control strategies are not broadly employed yet in clinical applications, mainly due to their poor ability to compensate for the variability over time of EMG signals caused by muscle fatigue, limb position effect, sweat, and electrodes shift \cite{kyranou_causes_2018}.
To address these issues, in this work we present three primary contributions:
\begin{itemize}
    \item \textbf{HD-sEMG setup:} we introduce an innovative HD-sEMG acquisition setup featuring 64 dry electrodes evenly distributed across the surface of the forearm. This setup aims to increase the amount of captured information while ensuring comfortable wearability, facilitated by a fully dry acquisition interface;
    \revtext{\item \textbf{Incremental HD-sEMG classification pipeline:} we propose an efficient and accurate incremental learning pipeline to address distribution shifts in HD-sEMG data over multiple days and achieve adaptive prosthesis myocontrol.}     
    Our proposed solution targets everyday prosthesis usage, 
    where subjects need to swiftly adapt the control model with limited data;
    \item \textbf{DELTA dataset: } DELTA is a unique new benchmark for long-term incremental learning on HD-sEMG data.
    We release the data acquired from 7 subjects, including \revtext{one with limb difference}, over 6 sessions across almost 4 months along with the code used in our experiments.
        
\end{itemize}
\IEEEpubidadjcol 
\section{Related works}
\noindent \textbf{Machine learning for prosthetic myocontrol: }
among the relevant applications of supervised learning for myocontrol, we recall the early study on amputees controlling a 
prosthesis via pattern recognition and outperforming conventional myocontrol during clinical tests~\cite{hargrove_pattern_2013}. 
\revtext{Enhancing myocontroller robustness and adaptability is a longstanding goal.
Data-efficient approaches
involving 
humans annotators result in improved adaptation with minimal data~\cite{szymaniak_recalibration_2022}.
Another well-known yet computationally expensive strategy is joint training across acquisition sessions \cite{sensinger_adaptive_2009}.
Supervised adaptation methods update model parameters on few samples of different data distributions \cite{chen_application_2013, vidovic_improving_2015}, 
while in~\cite{tommasi_improving_2013} model pre-training across subjects enables data-efficient adaptation.}
Recently, unsupervised learning has been employed
to facilitate 
co-adaptation between the controller and the user~\cite{gigli_progressive_2023}.
Moreover,
incremental learning 
approaches
update
models
on streams of data
to
improve 
accuracy
on
related tasks (class/task incremental)~\cite{nowak_simultaneous_2023} 
or tackle the distribution shift
of EMG data (domain incremental)~\cite{egle_preliminary_2023}.
Domain incremental methods 
proved effective
in predicting finger forces from sEMG signals to teleoperate a robotic arm~\cite{gijsberts_incremental_2011}, as well as for the on-edge deployment 
of classifiers
for gesture recognition \cite{burrello_tackling_2021}.
While tackling sEMG distribution shift, previous works only considered short time frames disregarding long-term changes occurring in real-world conditions. 
This work specifically targets the long-term stability of myocontrol systems tailored to gesture recognition in prosthetic applications.\\

\noindent\textbf{High-Density sEMG sensing: } several HD-sEMG acquisition systems have recently been presented
involving multiple sEMG electrodes positioned on a localized area of the body. 
In traditional HD-sEMG technology, conductive cream is commonly applied to minimize skin-electrode artifacts, especially when the A/D converter is distantly located from the acquisition point. Despite signal quality limitations, a dry acquisition bracelet has been developed for convenient use and wearability~\cite{tam_wearable_2019}, 
while employing data augmentation 
to provide shift invariance~\cite{f_chamberland_novel_2023}.
Other 
solutions include lightweight and screen-printed HD-sEMG arrays both with wet~\cite{moin_wearable_2021} and dry~\cite{murciego_novel_2023} skin-electrode interface. 
We introduce a dry HD-sEMG 
personalized
acquisition interface featuring 64 sEMG electrodes equally  distributed over the target surface. 
The dry interface improves 
wearability, making it easier to use.
Unlike previous works, our setup allows for the integration of the liner into an actual prosthesis (Fig.~\ref{fig:control_scheme}).\\

\noindent\textbf{Open myocontrol datasets: }
in~\cite{atzori_electromyography_2014}, an early effort to publish low and medium-density EMG data 
is presented. 
Several HD-sEMG datasets are also available in the literature, as summarized in Table \ref{tab:literature_datasets}. SEEDS~\cite{matran-fernandez_seeds_2019}, EMaGer~\cite{f_chamberland_novel_2023}, CapgMyo~\cite{du_surface_2017}, and Hyser~\cite{jiang_open_2021} encompass multiple subjects and gestures, 
yet fail to address temporal shift.
csl-hdemg~\cite{amma_advancing_2015}, includes data acquired over 5 sessions 
in which the wet HD-sEMG array is placed in slightly different positions.
In contrast with previous works, our DELTA dataset includes long-term HD-sEMG acquisition periods
and include a prosthetic end-user.

\begin{table}[!t]
    \caption{HD-sEMG datasets comparison.\label{tab:literature_datasets}}
    \centering
    \resizebox{\columnwidth}{!}{%
    \setlength\extrarowheight{1pt}
    \begin{tabular}{lccccc|c}
        \toprule
        \diagbox{\textit{Properties}}{\textit{Dataset}} & \thead{\textbf{SEEDS}\\ \cite{matran-fernandez_seeds_2019}} & \thead{\textbf{EMaGer}\\ \cite{f_chamberland_novel_2023}} & \thead{\textbf{CapgMyo}\\ \cite{du_surface_2017}} & \thead{\textbf{Hyser}\\ \cite{jiang_open_2021}} & \thead{\textbf{csl-hdemg}\\ \cite{amma_advancing_2015}} & \thead{\textbf{DELTA}\\ \textbf{(ours)}}\\
        \midrule
        \rowcolor{black!10}\textbf{\# Subjects} & 25 & 12 & 10-18 & 20 & 5 & 7\\
        \textbf{Amputees} & \ding{55} & \ding{55} & \ding{55} & \ding{55} & \ding{55} & \ding{51} \\
        \rowcolor{black!10}\textbf{\# Sessions} & 1 & 2 & 2 & 2 & 5 & 6\\
        \textbf{\# Repetitions} & 6 & 10 & 10 & 2 & 10 & 10\\
        \rowcolor{black!10}\textbf{\# Gestures} & 13 & 6 & 8-12 & 34 & 27 & 7\\
        \textbf{\# Electrodes} & 126 (+8) & 64 & 128 & 256 & 192 & 64\\
        \rowcolor{black!10}\textbf{Type} & wet & dry & wet & wet & wet & dry\\
        \textbf{Frequency [Hz]} & 2048 & 1000 & 1000 & 2048 & 2048 & 2000 \\
        \bottomrule
    \end{tabular}
    }
\end{table}
\section{Background}
\label{sec:Background}
In this Section, we outline the ML methods employed in this work. We focus on supervised classification, in which a classifier mapping input features to output classes is trained on labeled examples to minimize the expected loss.
We consider batch and incremental classifiers. In the batch setting, the classifier is trained on a fixed training set, while incremental methods progressively update it on a stream of examples.
\revtext{We conduct comparative experiments between 7 
batch classification methods: k-Nearest Neighbors (kNN), Linear Discriminant Analysis (LDA), Multi Layer Perceptron (MLP), Random Forest (RFor), Random Features Support Vector Machine (RF-SVM),}
Regularized Least Squares for Classification (RLSC~\cite{rifkin_regularized_2003}), Random Features RLSC (RF-RLSC), 
and the incremental variants of 
(RF-)RLSC.
We include both linear and nonlinear methods to determine whether the greater approximation capacity of nonlinear classifiers results in higher accuracy, especially under the large domain shift occurring in long-term HD-sEMG 
data.
We now 
recall the key elements of RLSC and RF-RLSC in their batch and incremental formulations.

\noindent\textbf{RLSC: } consider a training set $D=\{x_i,y_i\}_{i=1}^{n}$ with inputs $x_i \in {\mathcal{X} \subseteq \mathbb{R}^{m}}$ consisting of $m$ sEMG channel measurements  and  class labels $y_i \in \mathcal{Y} = \{1, ...,  c\}$ encoding the associated desired movements. 
RLSC consists of a linear model 
with  weight matrix $W \in \mathbb{R}^{m \times c}$.
Since RLSC employs the squared loss, the optimal weights $\widehat{W}$ minimizing the regularized empirical risk can be computed in closed form as: 
\begin{equation}
    \widehat{W} = (X^{\top}X+\lambda{I}_{m})^{-1}X^{\top}Y,
    \label{eq:RLSC}
\end{equation}
where $X \in \mathbb{R}^{n \times m}$ and $Y \in \mathbb{R}^{n \times c}$ are the matrices of stacked input ($x_i$) and output ($y_i$) vectors, respectively.
\revtext{The batch training computational cost is $O(n \cdot m^{2} + m^{3})$.
}
The predicted label $\hat{y} \in \mathcal{Y}$  for a new input $x\in \mathcal{X}$ can be computed as 
\begin{equation}
    \hat{y} = f(x) = {\text{argmax}} \{{\widehat{W}}^{\top}x\}.
    \label{eq:RLSCpred}
\end{equation}

\begin{figure*}[t]
\begin{minipage}[b]{0.36\linewidth}

\begin{algorithm}[H]
\caption{Incremental RLSC Training.}
\begin{algorithmic}
\STATE \textbf{Input:} $\lambda > 0$
\STATE \textbf{Output:} $W_{k}$
\STATE \textbf{Initialize:} $R_0 \gets \sqrt{\lambda}\cdot I_{m}$,
\STATE \hspace{1.6cm}$b_0 \gets \mathbf{0}$ with $b_0 \in \mathbb{R}^{m \times c}$,
\STATE \hspace{1.6cm}$W_0 \gets \mathbf{0}$ with $W_0 \in \mathbb{R}^{m \times c}$.
\STATE \textbf{Increment:}
\STATE \hspace{0.5cm} $b_{k} \gets b_{k-1} + x_{k}^\top \mathbf{e}_{y_{k}} $
\STATE \hspace{0.5cm} $R_{k}=$ \textsc{CholeskyUpdate}$(R_{k-1},x_{k})$
\STATE \hspace{0.5cm} $W_{k}=R_{k}^{-1}(R_{k}^\top)^{-1} b_k$
\STATE \textbf{return} $W_{k}$
\end{algorithmic}
\label{alg:incr_RLSC}
\end{algorithm}

\vspace{-0.3cm}
\end{minipage}
\hfill
\begin{minipage}[b]{0.63\linewidth}

    \centering
    \includegraphics[width=1.0\linewidth]{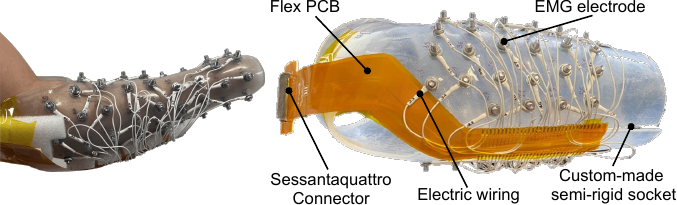} 

\captionof{figure}{Custom HD-sEMG socket for a limb difference subject (left) and for an able-bodied one (right). The 64 EMG electrodes are wired to flexible PCB with an integrated Sessantaquattro connector.\label{fig:socket}}
\end{minipage}
\end{figure*}

\noindent\textbf{Incremental RLSC: } incremental  learning algorithms sequentially update predictive models with new training samples and can adapt them to shifting distributions.
In particular, the closed-form RLSC solution shown in Eq.~\ref{eq:RLSC} can be conveniently
computed
in an incremental way~\cite{camoriano_incremental_2017}.
The updated model $W_k$ at step $k$ can be efficiently and exactly obtained by combining $W_{k-1}$ with the newly observed example $(x_k, y_k)$, as detailed in Alg. \ref{alg:incr_RLSC}.
\revtext{The computational cost of a single classifier update  is $O(m^2)$, while the full training procedure costs $O(n\cdot m^2)$ for $n$ updates.}
Model predictions at step $k$ can still be computed as in Eq.~\ref{eq:RLSCpred}, based on $W_k$.\\
\noindent\textbf{RF-RLSC (batch and incremental): } real-world learning problems often benefit from nonlinear modeling, and HD-sEMG classification is no exception.
RLSC models can be extended to the nonlinear case while retaining their incremental formulation thanks to kernel approximation schemes based on random sampling.
In particular, we consider Random Features \cite{rahimi_random_2007} due to their versatility and modularity. 
This involves projecting the input data into a feature space of typically higher dimensionality $M > m$ based on a nonlinear random feature map $\phi_M(x): \mathcal{X}\subseteq \mathbb{R}^m \rightarrow \mathcal{\tilde{X}}\subseteq \mathbb{R}^M$ approximating a  kernel function $k(\cdot,\cdot): \mathcal{X} \times \mathcal{X} \rightarrow \mathbb{R}$, i.e., $\langle\phi_M(x),\phi_M(x')\rangle  	\approx k(x,x')$.
Data mapped according to $\phi_M$ can be provided as input to batch or incremental RLSC models as presented earlier in this section, yielding efficient nonlinear classifiers with respect to the original input space.

\section{Materials and Methods}

\subsection{Acquisition Setup}
\label{sec:AcquisitionSetup}
We now describe the proposed acquisition setup in charge of measuring the forearm muscle activity while 
a set of 
movements
is performed. 
It consists of a novel 
HD-sEMG
interface based on dry electrodes placed on a semi-rigid custom liner. 
The motivation
for developing 
our acquisition setup is to enable \revtext{limb absence subjects} to use the 
prosthesis 
in their 
daily activities 
through a device which is easy to wear and operate. 
The 
main
limitation of commercial HD-sEMG systems lies in
their
requirement for effective skin-electrode contact, currently restricting their application to laboratory settings. 
This constraint arises from the necessity for conductive gel to reduce interface impedance. In our pursuit of a more user-friendly device, we have opted to transition from a wet 
interface to a dry one. 
Although this design choice may 
slightly decrease
signal quality, it significantly enhances 
usability
for prosthesis users outside the laboratory environment.
Moreover, with our 
innovative system, we aim at increasing the amount of collected data, thus acquiring information from most of the forearm muscles.
Therefore, the positioning of 64 electrodes is studied to cover the whole interested area while avoiding the bony zone.
Electrodes 1 to 32 are placed on the flexor muscles, whereas 33 to 64 are located on the extensor ones.
Since the sockets are individually customized, it is difficult to maintain uniformity in electrodes placement across subjects.
However, we aim to obtain a functional device for each individual, resulting in custom fit models.
This goal suits well prostheses end-users, since the varied muscle organization among \revtext{limb difference subjects} poses challenges for generalizing across this population.
The system shown in Fig.~\ref{fig:socket} (right) is the internal liner built for an able-bodied subject (patent application: 10202300002816). 
The personalized semi-rigid socket is the result of a novel manufacturing process developed in collaboration with 
INAIL Prosthetic Center in Budrio (Italy). 
The negative plaster counterpart is made from the cast of the subject's forearm. Subsequently, based on this mold, the internal thermolyn liner is formed. 
This ensures that each individual receives a personalized semi-rigid socket. 
The compliance of the liner enables to both conform to the forearm (or stump) and accommodate for volume changes during muscle contraction, while its rigidity ensures repeatable electrode placement on the skin over multiple dressings.
This approach enhances skin-electrode contact for improved functionality.
The electrodes consist of rigid conductive dome elements uniformly distributed on the liner with an inter-distance of approximately 2~cm. All electrodes are wired to a local flex printed circuit board (PCB), providing a compact connector for the HD-sEMG wireless acquisition system (Sessantaquattro, OT Bioelettronica, Turin, Italy).
The Sessantaquattro is a portable, multichannel (64 sensors) amplifier for recording of sEMG activity. The EMG signals are sampled at 2~kHz and collected in a monopolar configuration with respect to a reference electrode, fixed to the acromion through medical adhesive tape. The EMGs are A/D converted with 16~bits of resolution and transferred to a PC via Wi-Fi for
visualization and storage.

\subsection{The DELTA Dataset}
\label{sec:Datasets}
We introduce and release\footnote{The \textit{DELTA} dataset is available on Zenodo: \href{https://doi.org/10.5281/zenodo.10801000}{10.5281/zenodo.10801000}.} the \textbf{DELTA} dataset (\textbf{D}ense \textbf{E}lectromyography for \textbf{L}ong-\textbf{T}erm \textbf{A}daptive Control)
consisting of HD-sEMG data collected from 7 subjects over an extended time span. 
\revtext{\textit{DELTA} is a substantial new benchmark
for developing and evaluating data-driven models under conditions that closely align with real-world prosthetic applications (Tab.~\ref{tab:literature_datasets}). The dry electrode-skin interface simplifies the donning and doffing process compared to commercial HD-sEMG systems, allowing users to integrate it into their daily prosthetic use (Fig.~\ref{fig:socket}).}
By focusing on the temporal shift of HD-sEMG data, \textit{DELTA} fosters the development of robust and adaptive ML models, which could significantly enhance the efficacy of prosthetic systems.
The data collection is based on our novel acquisition setup (Sec.~\ref{sec:AcquisitionSetup}) equipped with a dense distribution of EMG sensors. 
We measure muscle activity over a long-term period ($3.9 \pm 2.2$ months) to observe data distribution shift over time. 
We recruited 6 healthy individuals (5 males), right handed, without known neuromuscular impairments, and aged between 25 and 34 ($28.5 \pm 3.0$ years).
Moreover, a subject (28yo) with congenital upper limb amputation is included in the study.
Written informed consent was obtained from participants before data collection. All experiments were conducted in line with the Declaration of Helsinki and approved by the ethical committee of Regione Liguria, Italy (Ref.: IIT\_REHAB\_HT01).
The collection spans 6 days of HD-sEMG acquisitions with at least a week's break between two consecutive sessions.
The days elapsed between the first and last session of acquisition for each subject range from 48 to 262. 
\revtext{Such extended duration is pivotal to encompass the diverse factors contributing to variability in the EMG data at multiple time scales, including electrode displacement, muscle fatigue, sweating, changes in volume, and added weight \cite{kyranou_causes_2018}.} 
The participants are instructed to execute 7 distinct gestures, each labeled accordingly, \revtext{corresponding to the available movements on the Hannes prosthesis~\cite{boccardo_development_2023, laffranchi_hannes_2020}}, i.e., hand closing (HC, 1), opening (HO, 2), wrist pronation (WP, 3), supination (WS, 4), wrist flexion (WF, 5), extension (WE, 6) and resting (Rest, 0). 
Acquisition begins as the subject starts executing the movement and persists during the steady state.
Throughout this phase, the subject is asked to keep contracting the muscles until the end of the $2~s$ acquisition period. 
After each iteration, the subject relaxes the muscles to prevent fatigue. 
This procedure is repeated 10 times for each gesture. 
On every acquisition day, the dataset is saved as an $n \times (m+1)$ matrix, where $n$ is the product between sampling frequency ($f_s=2~kHz$), number of repetitions ($n_{reps}=10$), time window of the acquisition ($t_{win}=2~s$), and the number of recorded gestures ($c=7$), while $m$ represents the 64 HD-sEMG signals. The last column contains class labels.
In the following, we refer to the training set as $\mathcal{D}_{d,s}^{tr}$, to the incremental update set as $\mathcal{D}_{d,s}^{up}$, and to the test set as $\mathcal{D}_{d,s}^{ts}$, where $d$ corresponds to the acquisition day (i.e., $d \in \left\{1, \ldots, 6\right\}$) and $s$ corresponds to the subject number (i.e., $s \in \left\{1, \ldots, 7\right\}$, where $s=7$ represents the \revtext{limb difference one}). 
The construction of these sets is different on the basis of the learning configuration (Fig.~\ref{fig:dataset_split}). 
See Sec.~\ref{sec:LearningPipeline} for dataset splitting and Sec.~\ref{sec:results_DataVisualization} for dataset visualization.

\subsection{Learning pipeline}
\label{sec:LearningPipeline}
\begin{figure}[t]
    \centering
    \includegraphics[width=0.9\linewidth]{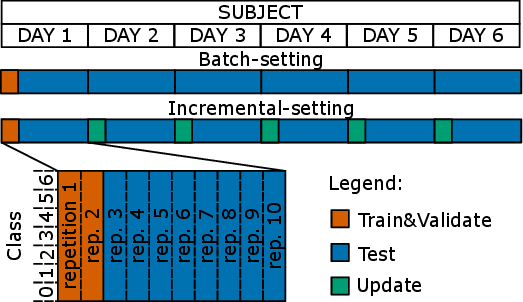} 
    \caption{DELTA dataset splits.
    Representation of 6 days of HD-sEMG acquisitions from a subject. 
    10 repetitions of each class are acquired each day. 
    \textit{Batch setting:} 2 repetitions for each class are included in the training set ($\mathcal{D}_{d=1,s}^{tr}$) for the first day, while the remaining constitute the test set.
    \textit{Incremental setting:} same split  for the first day, while for the following days 
    2 repetitions for each class are employed for incremental updates ($\mathcal{D}_{d,s}^{up}$) and the remaining 8 form the test set.}
    \label{fig:dataset_split}
\end{figure}

In this Section, we present the pipeline employed to investigate classifiers performance on the \textit{DELTA} HD-sEMG data 
provided by our acquisition setup.
The learning process is performed individually for each subject, thus obtaining subject-specific models. 
We opt for this approach since electrode placements on custom sockets vary across subjects. 
Indeed, developing a single model for all subjects proved too challenging given the difficulty in guaranteeing consistent electrode placement on identical muscles across individuals, especially with limb differences.
Two learning settings are compared, namely the \textit{batch setting} and the \textit{incremental setting}:

\noindent\textbf{Batch setting}: 
The model is trained from scratch using data from the first day and remains unchanged across the following ones (Fig.~\ref{fig:dataset_split}). 
2 gestures repetitions from the first day form the training set $\mathcal{D}_{d=1,s}^{tr}$ while the remaining 8 are included in the test set  $\mathcal{D}_{d=1,s}^{ts}$. 
Model selection is performed via 10-fold cross-validation (CV) on $\mathcal{D}_{d=1,s}^{tr}$. 
Finally, we include  the data of the remaining days ($d=2, \ldots, 6$) in the test set $\mathcal{D}_{d,s}^{ts}$.

\noindent\textbf{Incremental setting}: 
The model is first trained from scratch on $\mathcal{D}_{d=1,s}^{tr}$ including $20\%$ of the data from the first day.
It is then updated using 2 repetitions for each of the following days (see Fig.~\ref{fig:dataset_split}). 
For RLSC and RF-RLSC, this involves updating the weight matrix ($W_k$ in Alg.~\ref{alg:incr_RLSC}).
The  model is 10-fold cross-validated on $\mathcal{D}_{d=1,s}^{tr}$. Since we acquired 10 repetitions for each gesture, 2 of them are included in $\mathcal{D}_{d=1,s}^{tr}$, while the others are added to $\mathcal{D}_{d=1,s}^{ts}$.
In contrast with the \textit{batch setting}, in this case for each of the following days ($d=2, \ldots, 6$) we split the data into $\mathcal{D}_{d,s}^{up}$ ($20\%$) and $\mathcal{D}_{d,s}^{ts}$ ($80\%$).
The model is incrementally updated on 2 repetitions for each day ($d=2, \ldots, 6$), and sequentially tested on the remaining 8 repetitions.

In the following, we present the detailed description of each step of the learning pipeline: 
\subsubsection{Filtering}
The acquisition setup collects raw myoelectric data (spectrum range: 10~Hz to 500~Hz \cite{merletti_technology_2009}). 
To filter out noise,
we first apply a notch filter to remove the powerline frequency (50~Hz) followed by a Butterworth bandpass filter (20~Hz to 500~Hz) to remove motion artifacts.

\subsubsection{Preprocessing}
\label{sec:methods_LearnPipe_Process}
We employ the well-known Root Mean Square (RMS) 
representation:
$RMS(x) = \sqrt{1/N\sum_{i} x_{i}^{2}}$. 
Filtered EMG signals are used to compute RMS separately for each channel.
We use a sliding window of 200~ms ($N=400$ samples) and a 50~ms increment, in line with the acceptable controller delays for prosthetic applications~\cite{englehart_robust_2003, smith_determining_2010}. \revtext{To simplify the ML-model design, we fixed some parameters early, though experimenting with smaller window sizes~\cite{khushaba_decoding_2021} might slightly affect decoding performance. Moreover, different handcrafted features can impact the classifiers outputs~\cite{tigrini_intelligent_2024}.}
\subsubsection{Normalization}
\label{sec:methods_LearnPipe_Normaliz}
The RMS of the EMG signals are then normalized by subtracting the minimum value and dividing by the difference between the maximum and minimum of each of the 64 channels,
computed for each day. 

\subsubsection{Model Selection}
\label{sec:methods_LearnPipe_ModelSelect}
As for 
model training,
we consider
a realistic scenario in which a well-performing model shall be obtained from a limited budget of labeled data. 
Model selection follows the same procedure in the batch and incremental settings.
Specifically, it is carried out on $\mathcal{D}_{d=1,s}^{tr}$ 
(\textit{Train\&Validate} in Fig.~\ref{fig:dataset_split}). We perform a 10-fold cross-validation to select the best hyperparameters.
After determining the best model, it is fixed and employed
in the 
training and updating phases. 
Further details are available in Sec.~\ref{sec:results_ModelSelect}.
 
\subsubsection{Model Training and Update}
\label{sec:methods_LearnPipe_ModelTrain}
The 
classification methods
used in this 
work
are presented in Sec.~\ref{sec:Background}. 
After model selection, the optimal hyperparameters are 
fixed, paving the way for the training process. 
In this work, we compare incremental methods with batch 
ones, 
evaluating their adaptability to data distribution shifts over multi-day acquisitions.
In both settings,
training is performed on 
$\mathcal{D}_{d=1,s}^{tr}$.
Moreover, in the incremental setting, the update phase (\textit{Update} in Fig.~\ref{fig:dataset_split}) takes place on $\mathcal{D}_{d,s}^{up}$ for all the following days ($d=\{2, \ldots, 6\}$).

\subsubsection{Classification}
\label{sec:methods_LearnPipe_Classification}
In the testing phase, trained models are employed to classify previously unseen examples from $\mathcal{D}_{d,s}^{ts}$. 
Classification accuracy 
is computed for
the 8 test repetitions 
of each day.
\section{Experimental Results}
We now present the results obtained by the pipeline introduced in Sec.~\ref{sec:LearningPipeline} on the DELTA dataset, focusing on long-term adaptation. 
\revtext{We conduct the analysis on a Dell Precision workstation with a $12^{th}$ generation Intel Core i7 CPU.}
The experimental code is available at \href{https://github.com/DarioDiDomenico/Incr_HDsEMG.git}{https://github.com/DarioDiDomenico/Incr\_HDsEMG}.

\subsection{Data Distribution Shift Visualization}
\label{sec:results_DataVisualization}

\begin{figure}[t]
\begin{subfigure}[t]{0.48\linewidth}
    \centering
    \includegraphics[width=1\textwidth]{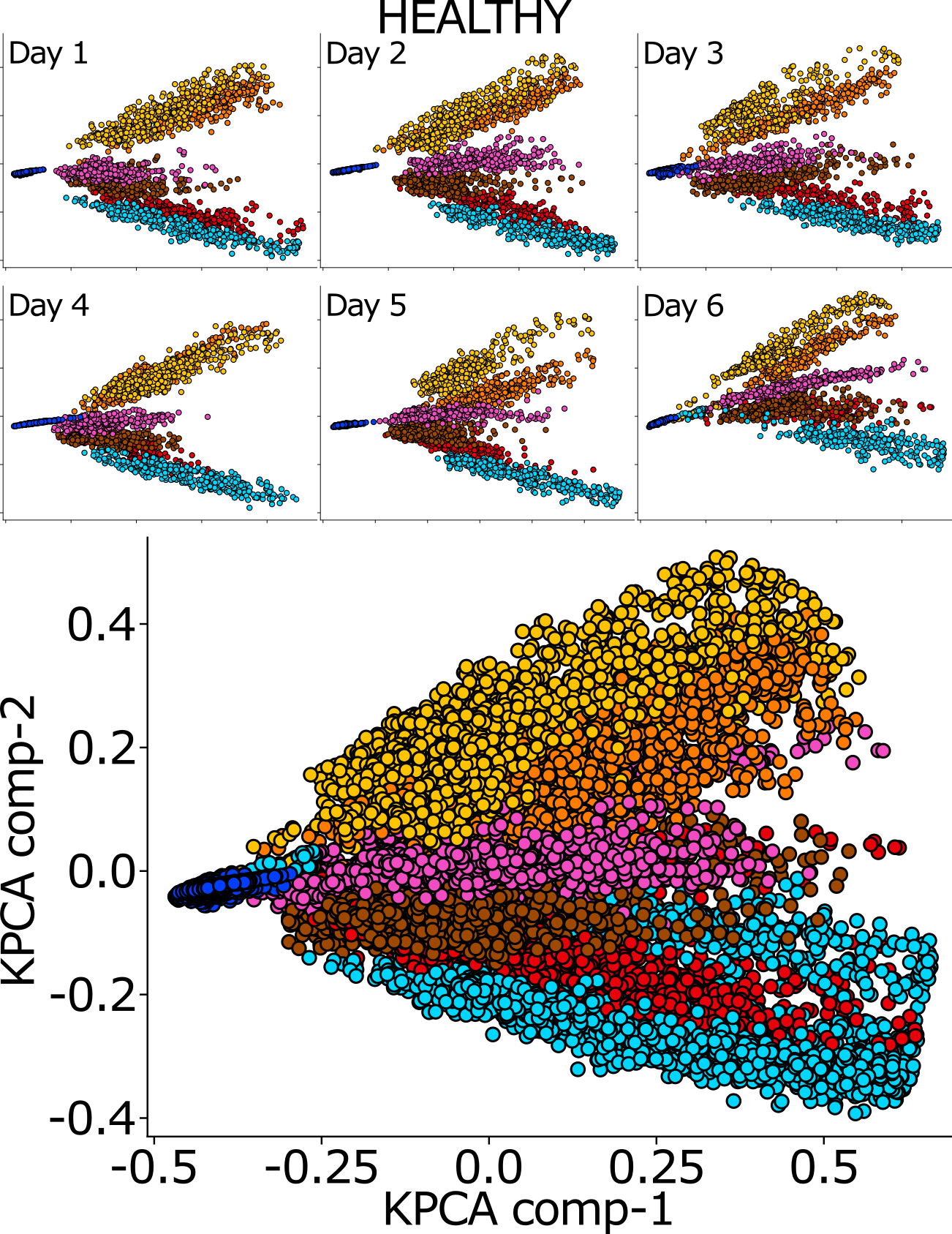}
    \caption{}
    \label{fig:KPCA_healthy}
\end{subfigure}
\hspace{0.2cm}
\begin{subfigure}[t]{0.48\linewidth} 
    \centering
    \includegraphics[width=1\textwidth]{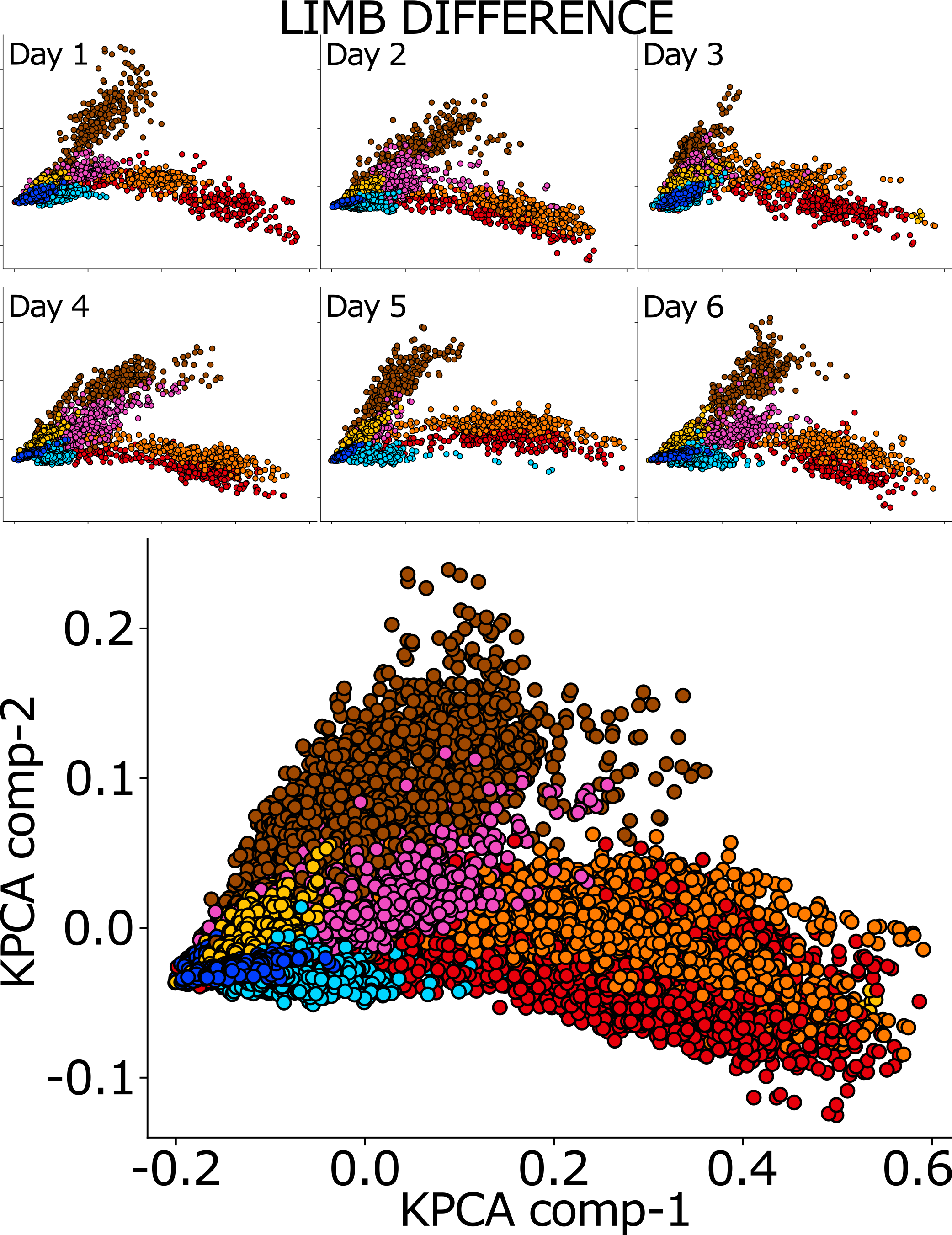}
    \caption{}
    \label{fig:KPCA_amputee}
\end{subfigure}
\caption{2D KPCA projection of the DELTA dataset  for each of the 6 acquisition days (top), as well as for the entire dataset across days (bottom), for (a)~a representative healthy subject and (b)~the \revtext{limb difference} subject. Colors represent classes.
}
\label{fig:KPCA}
\end{figure}

We first visualize the data distribution shift of the acquired HD-sEMG data across multiple days
by applying Kernel Principal Component Analysis (KPCA~\cite{scholkopf_nonlinear_1998}) with the Gaussian kernel to the full dataset and then plotting the data projected on the the first two KPCA nonlinear components.
We employ KPCA since it allows us to compute a single explicit nonlinear transformation of the data on the full dataset.
Then, we apply it to data from individual days to evaluate distribution shifts across days in the same 2D space.
We report the KPCA-projected data for a healthy subject (Fig.~\ref{fig:KPCA_healthy}, $\gamma_{KPCA}=0.05$) and the \revtext{limb difference one} (Fig.~\ref{fig:KPCA_amputee}, $\gamma_{KPCA}=0.01$). 
In Fig.~\ref{fig:KPCA}, the large plots include all the subject's data, while the small ones represent the data of each day. 
Note that data are relatively well-separable at single-day level, while overlap between classes on all days combined is
strong. Also, significant distribution shift across days can be observed for both subjects.
Such observations further motivate our investigation on the performance of batch and incremental classifiers 
on long-term HD-sEMG data 
to tackle distribution shift, and highlight the unique relevance of the DELTA dataset as a benchmark.

\subsection{Model Selection}
\label{sec:results_ModelSelect}

\begin{figure*}[t]
\begin{minipage}[b]{0.62\linewidth}

\captionof{table}{Optimal hyperparameters ($HP^{*}$) mean $\pm$ standard deviation for each participant. 
    Statistics computed over the model selection results for each day of HD-sEMG acquisition.
    S: healthy subject, LDS: \revtext{limb difference} subject.\label{tab:modSel2}}
\resizebox{\columnwidth}{!}{%
\centering
    \setlength\extrarowheight{2pt}
    \begin{tabular}{lc|ccccccc}
        \toprule
                                    \textit{Algorithm} &    $HP^{*}$    & \textbf{S1}          & \textbf{S2}          & \textbf{S3}          & \textbf{S4}          & \textbf{S5}          & \textbf{S6}          & \textbf{LDS1}          \\ 
        \midrule\rowcolor{black!10}
        \multicolumn{1}{l}{\cellcolor{white}\multirow{2}{*}{\textbf{RF-RLSC}}} & $\gamma^{*}$  & 0.2$\pm$0.2 & 0.4$\pm$0.2 & 0.5$\pm$0.2 & 0.4$\pm$0.4 & 1.0$\pm$1.3 & 0.2$\pm$0.2 & 0.2$\pm$0.3 \\
        \multicolumn{1}{l}{}       & $\lambda^{*}$ & 0.069$\pm$0.092 & 0.003$\pm$0.001 & 1.465$\pm$1.337 & 0.002$\pm$0.002 & 0.026$\pm$0.051 & 0.051$\pm$0.063 & 0.017$\pm$0.037 \\
        \rowcolor{black!10}
        \multicolumn{1}{l}{\cellcolor{white}\multirow{2}{*}{\textbf{RF-SVM}}} & $\gamma^{*}$  & 0.2$\pm$0.1 & 0.7$\pm$1.4 & 0.3$\pm$0.3 & 0.3$\pm$0.5 & 1.0$\pm$1.1 & 0.9$\pm$1.4 & 0.5$\pm$0.5 \\
        \multicolumn{1}{l}{}       & $C^{*}$  & (1.4$\pm$2.6)$\cdot 10^{3}$ & (2.1$\pm$2.6)$\cdot 10^{3}$ & 24.4$\pm$51.3 & (4.7$\pm$3.8)$\cdot 10^{3}$ & (3.3$\pm$3.4)$\cdot 10^{3}$ & (1.9$\pm$3.6)$\cdot 10^{3}$ & (2.4$\pm$2.1)$\cdot 10^{3}$ \\
         \rowcolor{black!10}
        \multicolumn{1}{l}{\cellcolor{white}\multirow{2}{*}{\textbf{MLP}}} & $l^{*}$  & 2.8$\pm$0.4 & 3.0$\pm$0.0 & 1.3$\pm$0.7 & 3.0$\pm$0.0 & 3.0$\pm$0.0 & 1.7$\pm$0.7 & 2.7$\pm$0.5 \\
        \multicolumn{1}{l}{}       & $h^{*}$  & 79.2$\pm$30.3 & 91.7$\pm$18.6 & 37.5$\pm$28.0 & 91.7$\pm$18.6 & 87.5$\pm$28.0 & 54.2$\pm$33.6 & 91.7$\pm$18.6 \\
        \rowcolor{black!10}\multicolumn{1}{l}{\textbf{kNN}}    & $k^{*}$      & 1.0$\pm$0.0 & 1.7$\pm$1.5 & 6.3$\pm$5.1 & 1.0$\pm$0.0 & 1.3$\pm$0.8 & 9.3$\pm$17.8 & 1.7$\pm$1.5 \\
        \multicolumn{1}{l}{\textbf{RLSC}}   & $\lambda^{*}$ & 1.5$\pm$0.8 & 31.7$\pm$39.4 & 5.7$\pm$3.8 & 0.4$\pm$0.7 & 1.5$\pm$2.6 & 3.1$\pm$3.0 & 1.3$\pm$1.7 \\
        \rowcolor{black!10}\multicolumn{1}{l}{\textbf{RFor}}   & $T^{*}$ & 57.7$\pm$37.0 & 60.7$\pm$33.0 & 46.3$\pm$31.4 & 78.8$\pm$33.1 & 46.5$\pm$10.0 & 55.2$\pm$24.0 & 62.2$\pm$28.2 \\
        \bottomrule
    \end{tabular}\label{tab:opt_hyperparams}
}

\vspace{-0cm}
\end{minipage}
\hfill
\begin{minipage}[b]{0.36\linewidth}

\centering
\includegraphics[]{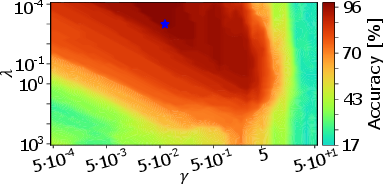}

\vspace{-0.2cm}
\captionof{figure}{$\lambda$ and $\gamma$ hyperparameters selection for RF-RLSC.
10-fold CV accuracy reported for a representative day. 
The color scale encodes the accuracy mean  minus 2 STD.\label{fig:RFRLSC_hyp_opt}}
\vspace{-0.3cm}
\end{minipage}
\end{figure*}
For all the learning methods (except LDA, which is hyperparameter-free), model selection is carried out on $\mathcal{D}_{d=1,s}^{tr}$ only, and the optimal hyperparameters remain constant while updating or testing the model on the remaining days. 
As detailed in Section \ref{sec:LearningPipeline}, hyperparameters 
are individually optimized for each subject as follows:

\noindent\textbf{kNN: } the number of neighbors $k$ is optimized on a range between 1 and 49.
We select the optimal $k^{*}$ exhibiting the highest predictive performance and robustness, defined as average accuracy minus two standard deviations (STD).

\noindent\revtext{\textbf{MLP: } we select the architecture considering up to 3 hidden layers and 25, 50, or 100 neurons.
The best one maximizes average accuracy minus 2 STD, prioritizing low model complexity first by layers ($l^{*}$), then by neurons ($h^{*}$).}

\noindent\revtext{\textbf{RFor: } we optimize the number of trees ($T$) in the forest across 50 values between 1 and 120. 
The optimal hyperparameters maximize
average accuracy minus 2 STD.}

\begin{figure}[t]
\includegraphics[width=\linewidth]{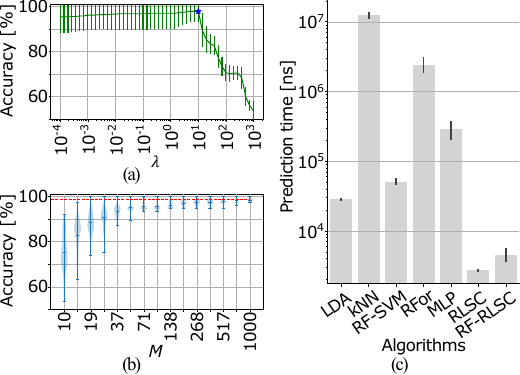}
\vspace{-0.55cm}
\caption{(a) RLSC: 
    CV accuracy
    vs. $\lambda$ for S4.
    Optimal $\lambda^*$ in blue.
    (b)~RF-RLSC: CV accuracy vs. RF dimensionality \textit{M} for S4.
    Upper bound in red. (c)~Classifier prediction times.
    }
\label{fig:opt_RLS_nComponents}
\end{figure}

\noindent\textbf{RLSC: } $\lambda$ is optimized over 50 values logarithmically scaled between $10^{-4}$ and $10^{3}$. $\lambda^{*}$ maximizes the average accuracy minus twice the STD. As shown in Fig.~\ref{fig:opt_RLS_nComponents}a for a representative subject, the accuracy gradually increases for growing values of $\lambda$, followed by a sharp decrease for values larger than the optimal $\lambda^*$.
This trend is observed across all subjects. 

\noindent\textbf{RF-RLSC: } 
we optimize 3 hyperparameters: the number random features $M$, the regularization parameter $\lambda$, and the Gaussian kernel coefficient $\gamma$. First, we select $M^*$ in a range $M \in \left\{10, \ldots, 1000\right\}$.
We report the validation accuracy with respect to $M$ in Fig.~\ref{fig:opt_RLS_nComponents}b, where the upper bound in red corresponds to the accuracy obtained with the exact kernel.
Due to the hardware constraints of the embedded control system, we select $M^*$ by trading off  accuracy and computational cost.
Across subjects, $M^*=500$ results in a low memory footprint and an accuracy consistent with the ideal one.
Then, we cross-validate $\lambda$ between $10^{-4}$ and $10^{3}$ and $\gamma$ in the range $ \left\{5\cdot10^{-4},\ldots, 50\right\}$,
considering 50 values in logarithmic scale for both. Fig.~\ref{fig:RFRLSC_hyp_opt} shows the validation performance for a representative subject. The color-scale represents the average accuracy minus twice the STD for each hyperaparameters combination. 
The optimal pair $(\lambda^{*}, \gamma^{*})$ is highlighted in blue.

\noindent\revtext{\textbf{RF-SVM: } we optimize the regularization parameter $C$ and the kernel coefficient $\gamma$.
The number of random features $M$ is fixed to 500, as detailed above. We cross-validate $C$ between $10^{-3}$ and $10^{4}$ and $\gamma$ in the range $ \left\{5\cdot10^{-4},\ldots, 50\right\}$, considering 50 values in logarithmic scale for both.}

\subsection{Batch vs. Incremental Methods Evaluation}
\label{sec:results_BatchVSIncr}

\begin{figure*}[t]
\centering
\includegraphics[]{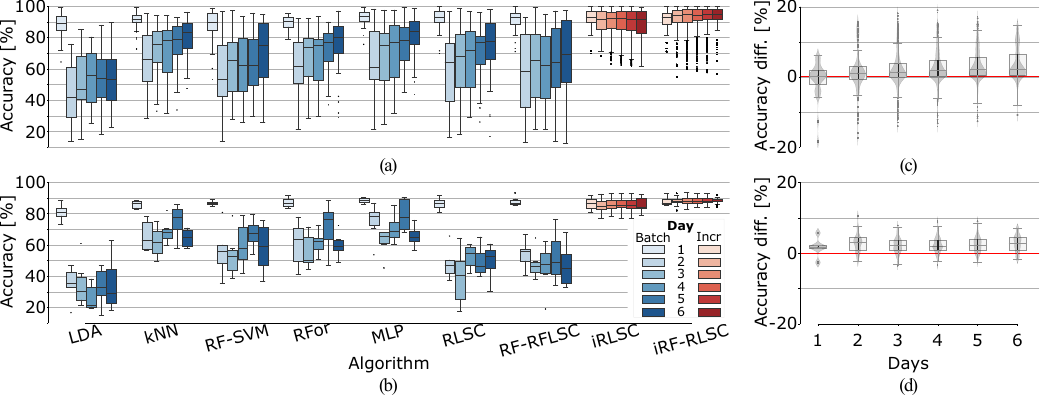}
\vspace{-0.12cm}
\caption{\textit{Batch} (blue) and \textit{incremental} (red) classifiers' accuracy over 6 days for  (a) 6 healthy subjects and (b) \revtext{limb difference one}.
    Distributions of the difference in test accuracy between incremental RF-RLSC and RLSC over 6 days for (c)~6 healthy subjects and (d)~\revtext{limb difference one}. The red line corresponds to equivalent accuracy (no difference).
        }
\label{fig:batchVSincr}
\end{figure*}

We evaluate and compare the classifiers in the \textit{batch} and \textit{incremental settings} by considering all possible permutations between days to exhaustively characterize the effects of a broad range of HD-sEMG distribution shifts on performance.
Given that model selection
is performed on $\mathcal{D}_{d=1,s}^{tr}$, the optimal hyperparameters only depend on the data of the first day in the specific permutation. Thus, since we consider all permutations, optimal hyperparameters potentially change
for each permutation, strengthening the generality and adherence to real-world conditions of our analysis.
In the \textit{batch setting}, the classifier is trained on the data of the first day ($\mathcal{D}_{d=1,s}^{tr}$) and  tested on all the remaining days.
In the \textit{incremental setting}, the model is trained on $\mathcal{D}_{d=1,s}^{tr}$ and incrementally updated on $\mathcal{D}_{d,s}^{up}$, with $d \in \{2,\ldots,6\}$.
Note that the classifier is updated on few examples (i.e., 2 repetitions, 28 seconds).
This reflects the scenario of day-to-day prosthesis use, in which the subject wears the device and the model quickly adapts.

Fig.~\ref{fig:batchVSincr} reports the accuracy of the classifiers on all  permutations for healthy subjects (Fig.~\ref{fig:batchVSincr}a) and the \revtext{limb difference one} (Fig.~\ref{fig:batchVSincr}b). We observe that \textit{batch} algorithms incur severe drops in accuracy on the remaining days due to their sensitivity to distribution shift.
The best-performing batch algorithm is kNN, whose accuracy drop is smaller, although very  significant (i.e.,~$ >10\% $).
In the \textit{incremental setting}, the results highlight high and stable accuracy across days for the incremental variants of (RF-)RLSC.
These trends are observed for both healthy subjects and the \revtext{limb difference one}.\\

\noindent\textbf{Impact of RF on incremental RLSC: }
We quantify the effect of random features on incremental (RF-)RLSC accuracy.
In particular, we evaluate the difference in test accuracy between  incremental  RF-RLSC and RLSC  in  the same experimental conditions on all subjects and days permutations.
Fig.~\ref{fig:batchVSincr}c-d displays the distributions of accuracy differences across days.
We observe that incremental RF-RLSC achieves consistently higher accuracy than linear RLSC for both subject categories, thus validating 
its effectiveness under distribution shift.
\section{Discussions}
As the number of prosthesis DoFs increases, so does the need for higher density and coverage of sEMG electrodes to accurately recognize a growing number of desired motions. However, myocontrollers incorporating HD-sEMG interfaces still lack wearability and ease of use.
In response to this challenge, we propose a novel system featuring 64 sEMG electrodes, thus enhancing information collection from the forearm, and a learning pipeline for classifying gestures from HD-sEMG data.
To the best of our knowledge, this is the first application of such a high number of dry EMG electrodes within a prosthetic liner, making it compatible with real prosthesis fittings.
Furthermore, we investigate the long-term performances of multiple learning methods
across multiple sessions. To this aim, we collect the DELTA dataset, composed of HD-sEMG data from 7 participants, including \revtext{one with limb absence}, over an extended period of nearly 4 months.
We observe strong distribution shift across days (Fig.~\ref{fig:KPCA}), which severely impacts classification performance.
In particular, classes are relatively well-separable within a single day, while the shift becomes evident across different days.
This causes accuracy drops when testing on new days models trained on previous ones, indicating the need to adapt the classifier to effectively adapt the controller. 
To address this issue, we employ incremental  classification methods and compare their predictive performance with multiple batch baselines. Our results show that incremental RLSC and RF-RLSC successfully adapt to long-term HD-sEMG shift, while batch ones suffer from significant performance degradation (Fig.~\ref{fig:batchVSincr}). 
\revtext{By including prosthetic end-users with congenital amputation, we demonstrate the system's accuracy and robustness over nearly 4 months of HD-sEMG acquisitions, despite challenging data distribution shifts.} Additionally, results in Fig.~\ref{fig:batchVSincr}c-d show that random features significantly improve accuracy in the incremental setting, demonstrating  the benefits of  incremental RF-RLSC in terms of accuracy and efficiency for HD-sEMG gesture recognition under distribution shift.
\revtext{
        Moreover, Fig.~\ref{fig:opt_RLS_nComponents}c reports prediction times for each classifier on the test set across all subjects and permutations. 
        Crucially, both RLSC and RF-RLSC display 10- to 1000-fold faster predictions than the other models, which is highly suitable for the hardware and time constraints of embedded prosthesis control systems. The decision to use (RF-)RLSC in the incremental setting is driven by their sustained accuracy over long-time, their efficacy with minimal hyperparameters, and their low prediction times. However, our study does not offer a comprehensive benchmark comparison of all available algorithms, especially in the incremental setting. Therefore,} 
we release the DELTA dataset as a unique benchmark for fostering the study of long-term dry HD-sEMG myocontrol.
\section{Conclusions}

We introduce a novel HD-sEMG customizable acquisition setup and an incremental learning pipeline for myocontrol enabling efficient and long-term adaptation with minimal labeled data. We release the DELTA HD-sEMG dataset collected from 7 participants, including \revtext{one with limb absence}, over an extended period of almost 4 months to stimulate research on this challenging task.
Our results show that fast (RF-)RLSC incremental updates result in high accuracy while satisfying tight memory and time constraints.
To our knowledge, this is the first study characterizing the performance of incremental learning methods for long-term, dry HD-sEMG myocontrol.

\bibliographystyle{IEEEtran}
\bibliography{root.bib}

\vfill
\end{document}